\documentclass{bmvc2k}
\usepackage{multirow}
\usepackage{comment}
\usepackage{amsfonts}
\usepackage{graphicx}

\title{
Sketch2TikZ: Multi-Reward Reinforcement Learning for Converting Hand-Drawn Sketches to Executable TikZ}

\addauthor{Shubham Paliwal}{shubham.p3@tcs.com}{1}
\addauthor{Arushi Jain}{j.arushi@tcs.com}{1}
\addauthor{Monika Sharma}{monika.sharma1@tcs.com}{1}
\addauthor{Manasi Patwardhan}{manasi.patwardhan@tcs.com}{2}

\addinstitution{
 TCS Research\\
 New Delhi\\
 India
}
\addinstitution{
 TCS Research\\
 Pune \\
 India
}

\runninghead{Paliwal et al}{Sketch2TikZ}

\begin{document}
\maketitle

\begin{abstract}
Automatic digitization of hand-drawn sketches into editable formats is crucial for modern document workflows and LaTeX integration. Converting sketches into executable TikZ code provides a structured representation of diagrams, but remains challenging due to noisy and ambiguous inputs, which hinder reliable interpretation and limit the effectiveness of token-level supervision for ensuring global diagram validity.
To address these challenges, we propose \emph{Sketch2TikZ}, a reinforcement learning framework that optimizes sketch-to-code generation via a sequential multi-reward strategy, combining Code-Structure ($\mathcal{R}_{cs}$), OCR ($\mathcal{R}_{ocr}$), CleanDIFT ($\mathcal{R}_{cd}$), and Self-Similarity ($\mathcal{R}_{ss}$) rewards to enforce textual, visual and structural correctness. We evaluate Sketch2TikZ on the \emph{SkeTikZ} and \emph{FlowChart-3B} benchmarks using metrics including compilation success rate (CSR), image similarity, code similarity, 
and character similarity. Our method achieves near-perfect compilation success and improves visual and textual alignment over strong open-source baselines, while remaining competitive with proprietary GPT-based vision–language models. Extensive ablations demonstrate that curriculum-based reward scheduling is critical for reducing reward interference, stabilizing training and mitigating proxy over-optimization. Overall, our results highlight the effectiveness of reference-aware, multi-objective reinforcement learning for structured diagram generation from noisy sketches.


\end{abstract}

\section{Introduction}
\label{sec:intro}
Scientific diagrams are central to effective scholarly communication, enabling the precise depiction of complex concepts, algorithmic workflows and system architectures. Among existing tools, TikZ has become the widely adopted framework for producing high-quality, publication-ready graphics within LaTeX documents~\cite{tantau2013tikz}. However, despite its expressiveness, authoring TikZ code manually remains a tedious and error-prone process. Users must manually translate visual intent into low-level programmatic constructs, often involving intricate coordinate specifications and styling rules. This steep learning curve creates a significant bottleneck in the document creation workflow.


Recent efforts have sought to alleviate this burden through automated diagram generation~\cite{saito2025sketch2diagram, wei2025diagramagent}. While text-to-code~\cite{zala2023diagrammergpt} approaches have shown some promise, describing complex spatial layouts and structural relationships using natural language is inherently cumbersome and often ambiguous~\cite{395661}. In contrast, sketch-based generation~\cite{bonnici2019sketch, sezgin2001sketch} offers a more natural and expressive alternative. Prior work~\cite{tan2025sketchagent, lee2023learning} has shown that hand-drawn sketches effectively capture geometry, layout, and textual content in a unified form, providing a direct bridge between human ideation and digital rendering.


However, converting such sketches into executable TikZ code remains highly challenging. Hand-drawn inputs are inherently noisy with ambiguous, incomplete, or overlapping strokes. Accurately recognising handwritten text embedded within geometric structures further complicates the task. Moreover, generative models must infer precise spatial relationships from imprecise drawings and map them to consistent coordinate systems. Further, TikZ code generation imposes strict syntactic and structural constraints; outputs must not only be visually faithful but also compile without errors. Existing vision-language models (VLMs)~\cite{wang2024llmcodefail, wu2023rustgen, chou2025vlmconsistency, yang2025qwen3} often struggle to satisfy these requirements simultaneously, leading to structurally inconsistent code, visual mismatches or compilation failures.

To overcome these limitations, we propose a novel framework based on Group Relative Policy Optimization (GRPO)~\cite{deepseek-math}. Instead of relying solely on standard supervised fine-tuning which often fails to enforce the intricate constraints of diagrammatic code generation, our approach leverages reinforcement learning to directly optimize the generative policy against task-specific objectives. At the core of our method is a sequential multi-reward design that provides targeted and interpretable feedback throughout the generation process. Instead of collapsing evaluation into a single objective, we decompose it into specialised reward functions, each addressing a distinct aspect of the task. This includes enforcing textual correctness, visual alignment, structural consistency and code validity in a progressive manner. Such staged optimization enables the model to refine different facets of generation incrementally, ultimately producing high-fidelity and syntactically reliable TikZ code.


To summarize, the key contributions of this work are as follows:
\begin{itemize}

\item We introduce \textbf{Sketch2TikZ}, a reinforcement learning framework based on Group Relative Policy Optimization (GRPO) for the challenging multi-modal sketch-to-TikZ generation task. Our approach employs a novel sequential multi-reward design that decomposes learning into specialised feedback signals: OCR reward for textual fidelity, CleanDIFT and Self-Similarity rewards for visual and spatial alignment, and Code-Structure reward for ensuring syntactic correctness and compilability.
\item We conduct comprehensive experiments demonstrating that our proposed Sketch2TikZ method outperforms existing open-source baselines with comparable parameters and achieves performance that is competitive with the state-of-the-art proprietary models like GPT-4o-mini.
\item We provide an in-depth ablation analysis detailing the individual and collective impact of the proposed reward components, offering valuable insights into optimal reward formulation for structurally constrained code generation tasks.
\end{itemize}

\begin{figure}
\includegraphics[width=\textwidth]{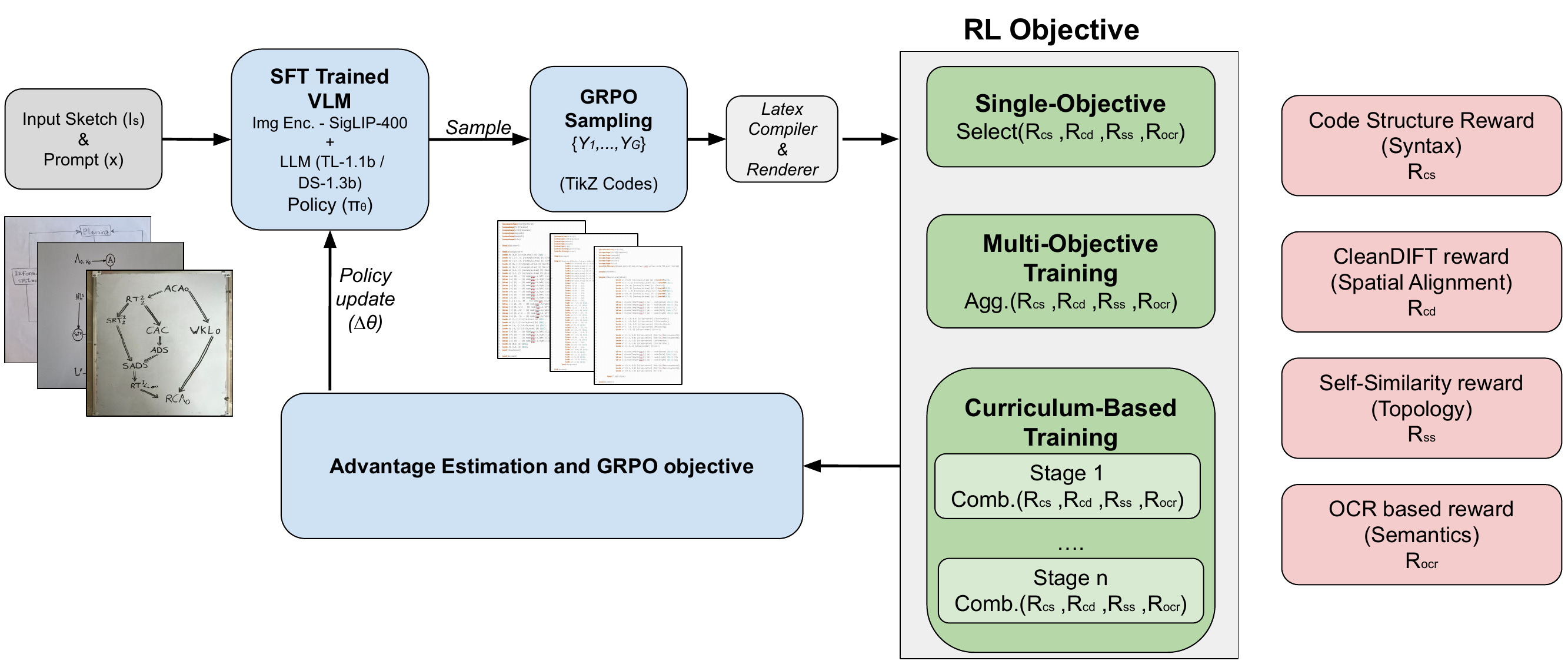}
\caption{Overview of the proposed RL optimization architecture under different settings.
A multimodal language model, initialized via supervised fine-tuning (SFT), translates hand-drawn sketches into TikZ code. The RL objective is formulated under three training configurations, as illustrated. The four proposed reward components, shown on the right, are combined in different ways across these configurations to guide RL-based optimization.}
\label{fig:architecture-overview}
\end{figure}

\section{Related Work}
Our work lies at the intersection of sketch-based diagram understanding and reinforcement learning for vision–language alignment.\\

\noindent \textbf{Sketch Understanding and Sketch-to-Diagram Translation}: Early approaches focused on sketch recognition and embedding learning~\cite{liu2017deepsketchhashingfast, yu2015sketchanetbeatshumans}. Subsequent works explored stroke-based generative modeling~\cite{das2020beziersketchgenerativemodelscalable, ha2017neuralrepresentationsketchdrawings} and perceptual grouping strategies~\cite{li2018universalperceptualgrouping}, which rely on temporal stroke information or heuristic grouping rules. However, these methods are not directly applicable to static, in-the-wild sketch images where stroke ordering is unavailable and visual ambiguity is significantly higher. More recently, \textit{Sketch2Diagram}~\cite{saito2025sketch2diagram} introduced the \textit{SkeTikZ} dataset and the \textit{ImgTikZ} model, establishing the first large-scale benchmark for mapping sketches directly to TikZ programs. Despite strong performance under supervised fine-tuning (SFT), the model is trained using Maximum Likelihood Estimation (MLE), which is known to suffer from exposure bias and teacher-forcing limitations~\cite{bengio2015scheduledsamplingsequenceprediction}. As a result, such approaches optimize token-level exact match rather than global structural correctness, often leading to invalid or non-compilable TikZ outputs and hallucinated textual content. In contrast, our method moves beyond token-level optimization to objective-driven reinforcement learning.\\

\noindent\textbf{Reinforcement Learning via Reference-Free Visual Rewards}:
Reinforcement learning (RL) has emerged as a powerful paradigm for aligning generative models with complex, non-differentiable objectives~\cite{mnih2013playing, schulman2017proximalpolicyoptimizationalgorithms}. In code generation, prior works such as PPOCoder~\cite{shojaee2023executionbasedcodegenerationusing} and CodeRL~\cite{le2022coderlmasteringcodegeneration} leverage execution-based rewards to guide generation, treating compilation or runtime correctness as supervision signals. More recent algorithms including Group Relative Policy Optimization (GRPO)~\cite{liu2026advancesgrpogenerationmodels}, improve the efficiency of RL-based training by eliminating the need for a learned value function via group-wise relative advantage estimation. However, extending such approaches to diagram synthesis introduces a critical challenge: defining appropriate reward signals. Conventional text-based metrics, such as BLEU~\cite{papineni-etal-2002-bleu}, are fundamentally misaligned with visual code generation tasks, as syntactically different programs may produce identical rendered diagrams. To address this limitation, recent efforts advocate for rendering-based and reference-free evaluation schemes~\cite{nguyen2026onlinereferencefreeevaluationframework}. Inspired by these approaches, we formulate diagram generation as a multi-objective optimization problem driven purely by visual and execution signals, without relying on ground-truth code matching.

\section{Methodology}

\subsection{Problem Formulation} 
Let \( I_s \in \mathbb{R}^{H \times W \times 3} \) denote an RGB hand-drawn sketch image. Our goal is to learn a mapping function \( f_{\theta} \), parameterised by \( \theta \), that transforms the input sketch into a structured textual representation in the form of TikZ code \( Y \):
\begin{equation}
Y = f_{\theta}(I_s, x),
\end{equation}
where $x$ is preset text instruction and \( Y = \{y_1, y_2, \dots, y_T\} \) is a sequence of tokens forming syntactically valid and compilable TikZ code.

The objective is to ensure that the generated code \( Y \) satisfies three key properties: (i) \textit{visual fidelity}, such that the rendered output closely matches the input sketch; (ii) \textit{textual correctness}, ensuring accurate recognition and reproduction of embedded textual elements; and (iii) \textit{structural validity}, guaranteeing syntactic correctness and logical consistency of the code.

To achieve this, we formulate \textbf{Sketch2TikZ} generation as a sequential decision-making problem, where the model iteratively generates tokens conditioned on the input sketch and previously generated tokens. We then optimize the policy \( f_{\theta} \) using reinforcement learning, enabling the incorporation of task-specific rewards that explicitly capture visual, textual, and structural objectives. We next describe the design and components of our proposed framework.

\subsection{Model Architecture} 
We adopt a vision-language architecture similar to prior works such as ImgTikZ~\cite{saito2025sketch2diagram} and LLaVA-1.5~\cite{liu2023llava}, comprising a frozen high-resolution visual encoder, a projection adapter, and an auto-regressive language model for code generation. Building upon this backbone, we explore two complementary architectural variants: (i) a \textit{general-purpose reasoning model} based on \textbf{TinyLlama-1.1B}~\cite{zhang2024tinyllama}, and (ii) a \textit{code-specialised model} based on \textbf{DeepSeek-Coder-1.3B}~\cite{guo2024deepseekcoder}, to analyse the trade-off between generic reasoning capabilities and domain-specific priors for structured code generation.
For visual feature extraction, both model variants adopt \textbf{Sigmoid Language-Image Pre-training} (SigLIP)~\cite{zhai2023siglip} as the backbone image encoder. Given an input sketch image \( I_s \in \mathbb{R}^{H \times W \times C} \), the encoder generates a sequence of localised visual embeddings that capture fine-grained spatial and structural cues present in the sketch.
\begin{equation}
    F_v = SigLIP(I_s) \in \mathbb{R}^{N\times D_v}
\end{equation}
where $N$ denotes the number of patch tokens and $D_v$ represents the visual embedding dimension.
The extracted visual features are subsequently passed through a Multi-Layer Perceptron (MLP) adapter to align them with the input embedding space of the language decoder. The adapter consists of a two-layer MLP with GELU~\cite{hendrycks2016gelu} activation, which projects the visual features from the visual embedding space $D_v$ to the textual embedding space $D_l$ of the language model.
Finally, we employ two lightweight language decoders, TinyLlama-1.1B and DeepSeek-Coder-1.3B, for structured code generation. Next, we train the model in two stages. The model is trained in a two-stage pipeline, as illustrated in Figure~\ref{fig:architecture-overview}. In the first stage, we perform supervised fine-tuning (SFT) on a paired SkeTikZ dataset~\cite{saito2025sketch2diagram}, enabling the model to learn an initial alignment between visual tokens and the syntactic patterns of LaTeX/TikZ, thereby establishing a reliable visual-to-text base policy. In the second stage, we further refine this policy using reinforcement learning via Group Relative Policy Optimization (GRPO)~\cite{deepseek-math}. Starting from the SFT-initialised policy, GRPO optimizes generation by sampling multiple candidate outputs per input prompt and updating the policy based on their relative rewards, thereby encouraging outputs that better satisfy task-specific objectives.

In this phase, we incorporate our proposed \textbf{sequential multi-reward framework}, consisting of \textbf{Code-Structure reward ($\mathcal{R}_{cs}$)}, \textbf{OCR-based Reward ($\mathcal{R}_{ocr}$)}, \textbf{CleanDIFT Reward ($\mathcal{R}_{cd}$)}, and \textbf{Self-Similarity reward ($\mathcal{R}_{ss}$)}. These rewards are designed to progressively enforce textual accuracy, visual fidelity, and structural correctness. Notably, GRPO enables direct optimization of the policy without requiring a separate critic network, allowing us to efficiently fine-tune the adapter and language decoder while maximising both compilability and visual alignment of the generated TikZ code.



\subsection{Multi-Reward Framework}
\label{sec:mrs} 

\subsubsection{Code-Structure Reward ($\mathcal{R}_{cs}$)} 
Unlike natural language, TikZ generation is governed by strict LaTeX compilation constraints, where even minor structural errors can lead to fatal failures. To provide informative supervision under such conditions, we introduce a two-tiered Code-Structure reward ($\mathcal{R}_{cs}$), designed to deliver dense feedback even when compilation fails.

We first employ \textit{ChkTeX}~\cite{chktex}, a lightweight static LaTeX analysis tool, to capture fundamental structural inconsistencies. Let $\mathcal{E}_{chk}(Y)$ denote the number of structural warnings identified by \textit{ChkTeX}, such as unclosed environments, mismatched brackets, or malformed commands. However, as \textit{ChkTeX} operates as a general-purpose text-based checker rather than a TikZ-specific parser, a zero-error report does not guarantee successful compilation. To address this limitation, we additionally evaluate binary compilation success $C(Y) \in \{0, 1\}$ using the \textit{pdflatex}~\cite{pdflatex} engine, which serves as definitive check for compilation success and executable validity of the generated graphical code. The reward is, therefore, structured hierarchically as follows:


\begin{equation} \mathcal{R}_{cs}(Y) = \alpha e^{-\tau \mathcal{E}_{\text{chk}}(Y)} + r_{\text{base}}\mathbf{1}[C(Y)=1] \end{equation}
where $r_{\text{base}}$ provides a strong positive reward upon successful compilation. The exponential decay term controlled by $\tau$ ensures that even non-compilable outputs receive a smooth and informative learning signal, which progressively improves as the policy reduces syntactic errors (i.e., $\mathcal{E}_{chk}(Y) \rightarrow 0$).


\subsubsection{OCR Reward ($\mathcal{R}_{ocr}$)} 
Hand-drawn sketches inherently combine spatial structures with embedded textual annotations (e.g., mathematical symbols and node labels). To penalise text hallucinations and localisation errors, we introduce OCR reward $\mathcal{R}_{ocr}$, which measures the fidelity of textual content preserved in the rendered output $\hat{I}_s = \text{Render}(Y)$ relative to the input sketch $I_s$.
Let $\mathcal{T} = \{(t_i, b_i)\}_{i=1}^{N}$ denote the set of text instances $t_i$ and their corresponding bounding boxes $b_i$, extracted from the input sketch $I_s$ after pre-processing using an off-the-shelf OCR engine. Similarly, let $\hat{\mathcal{T}} = \{(\hat{t}_j, \hat{b}_j)\}_{j=1}^{M}$ represent the text instances obtained from the rendered image $\hat{I}_s$.
We compute a bipartite matching between $\mathcal{T}$ and $\hat{\mathcal{T}}$ by jointly considering textual similarity and spatial overlap. Specifically, we measure textual fidelity using a normalised Levenshtein similarity defined as:
\begin{equation}
    \mathcal{S}_{str}(t_i, \hat{t}_j) = 1 - \frac{Levenshtein(t_i, \hat{t}_j)}{max(|t_i|, |\hat{t}_j|)} 
\end{equation}

For spatial fidelity, to account for minor layout distortions inherent to TikZ rendering, we employ a Soft Intersection-over-Union (Soft-IoU) metric, which provides a tolerant measure of overlap between predicted and ground-truth text regions.
\begin{equation}
    \mathcal{S}_{iou}(b_i, \hat{b}_j) = \frac{\mathcal{A}(b_i \cap \hat{b}_j)}{\mathcal{A}(b_i) + \mathcal{A}(\hat{b}_j) - \mathcal{A}(b_i \cap \hat{b}_j) + \epsilon}
\end{equation}
where $\mathcal{A}(\cdot)$ denotes the continuous area function. To handle cases where bounding boxes do not strictly overlap but remain spatially close, we introduce a Gaussian centre-distance penalty that provides a smooth and continuous measure of proximity. Let $c_i$ and $\hat{c}_j$ denote the center coordinates of $b_i$ and $\hat{b}_j$, respectively. By normalising the Euclidean distance with the image diagonal (assumed to be $1.0$ in normalised coordinates), we define the soft spatial signal using a Radial Basis Function (RBF) kernel~\cite{math12060829}:
\begin{equation}
    \mathcal{S}_{soft}(b_i, \hat{b}_j) = \exp \left( - \alpha_{ocr}|c_i-\hat{c}_j\|_2^2\right)
\end{equation}

While the standard IoU metric captures exact spatial overlap and is well-suited for evaluation, it is less reliable for hand-drawn sketches, where layout variations are common and text instances may differ in length or alignment. To address this ambiguity, we combine both the soft spatial similarity $\mathcal{S}_{soft}$ and IoU-based similarity $\mathcal{S}_{iou}$ using a max operation, thereby ensuring robustness to minor spatial deviations. The overall OCR reward is then computed by evaluating the joint similarity for each ground-truth text entity and determining optimal correspondences via the Hungarian matching algorithm. Defining $\pi^\star$ as the optimal one-to-one assignment obtained using the Hungarian algorithm, the reward is formalised as:
\begin{equation}
\mathcal{R}_{ocr}(I_s,\hat{I}_s)
=
\frac{1}{N}
\sum_{i=1}^{N}
\left[
\max\left(\mathcal{S}_{iou}(b_i,\hat{b}_{\pi^\star(i)}),
\mathcal{S}_{soft}(b_i,\hat{b}_{\pi^\star(i)})\right)
\cdot
\mathcal{S}_{str}(t_i,\hat{t}_{\pi^\star(i)})
\right].
\end{equation}

\subsubsection{Image Diffusion CleanDIFT Reward ($\mathcal{R}_{cd}$)} 
While $\mathcal{R}_{cs}$ ensures syntactic correctness and compilability, and $\mathcal{R}_{ocr}$ enforces textual accuracy, it is equally important to preserve the global geometric layout of the rendered diagram $\hat{I_s}$ in alignment with the input sketch $I_s$. Prior works (e.g., DeTikZify~\cite{belouadi2024detikzify}) typically rely on LPIPS-based perceptual similarity; however, this approach is sub-optimal due to the significant domain gap between noisy hand-drawn sketches and clean, vector-rendered outputs. To enforce stronger topological consistency, we leverage noise-free CleanDIFT~\cite{stracke2025cleandift} features, denoted as $F$ and $\hat{F}$, extracted from the sketch and the rendered image, respectively. Given a set of $M$ salient points $\mathcal{P} = \{p_i\}_{i=1}^{M}$ sampled from the input sketch, we establish cross-domain correspondences $\hat{\mathcal{P}} = \{\hat{p}_i\}_{i=1}^{M}$ within $\hat{I}_s$.


Since vector rendering inherently introduces unconstrained affine transformations (e.g., scaling and translation), direct coordinate matching becomes unstable. To address this, we design an affine-invariant reward based on $N$-nearest neighbour ($N$-NN) topological consistency. Let $\mathcal{N}_N(p_i)$ denote the index set of the $N$ spatial nearest neighbours of $p_i$ within $\mathcal{P}$, and let $\mathcal{N}_N(\hat{p}_i)$ denote the corresponding $N$-nearest neighbours of $\hat{p}_i$ within $\hat{\mathcal{P}}$. We then define the CleanDIFT reward $\mathcal{R}_{cd}$ as the mean neighbourhood overlap ratio:
\begin{equation}
    \mathcal{R}_{cd}(I_s,\hat{I_s}) = \frac{1}{M} \sum_{i=1}^{M} \frac{|\mathcal{N}_N(p_i) \cap \mathcal{N}_{N}(\hat{p}_i) |}{N}
\end{equation}

\subsubsection{Self-Similarity Reward ($\mathcal{R}_{ss}$)} 
Finally, to obtain a robust cross-domain visual reward, we project both $I_s$ and $\hat{I}_s$ into a shared semantic space using an ensemble of frozen vision foundation models $\Phi$ (e.g.,  CLIP~\cite{radford2021clip}, DINO~\cite{caron2021dino}, and OpenCLIP~\cite{cherti2023openclip}), which encode strong structural priors while remaining invariant to low-level texture variations. Let $\Phi_{m}^{(l)}(I) \in \mathbb{R}^{H_l \times W_l \times C}$ denote the dense spatial feature map extracted from layer $l$ from $m$ frozen model. We define $\mathcal{R}_{ss}$ as the spatially-aware cosine similarity between these dense feature representations. To account for potential misalignment introduced by scale and translation differences between the input sketch and the rendered output, we apply a local spatial max-pooling operation $\mathcal{P}_{k \times k}$ prior to computing the channel-wise cosine similarity.
\begin{equation} \begin{gathered} c_{m,h,w}^{(l)} = \frac{ \mathcal{P}(\Phi_m^{(l)}(I_s))_{h,w} \cdot \mathcal{P}(\Phi_m^{(l)}(\hat{I}_s))_{h,w} }{ \|\mathcal{P}(\Phi_m^{(l)}(I_s))_{h,w}\|_2 \, \|\mathcal{P}(\Phi_m^{(l)}(\hat{I}_s))_{h,w}\|_2 }, \\ \mathcal{R}_{ss}(I_s,\hat{I}_s) = \frac{1}{M} \sum_{m=1}^{M} \frac{1}{H_lW_l} \sum_{h,w} c_{m,h,w}^{(l)} \end{gathered} \end{equation}
This formulation improves upon standard LPIPS by explicitly disentangling sketch-specific noise (e.g., pencil textures, stroke irregularities) from the underlying structural content. As a result, the optimization objective focuses on preserving high-level geometric and topological consistency, rather than low-level appearance artefacts.

\section{Experiments}
In this section, we present a comprehensive evaluation of the proposed GRPO-based multi-reward Sketch2TikZ framework. We begin by describing the experimental setup, including the dataset, baseline models and evaluation metrics used in our experiments. We then provide a rigorous quantitative comparison with both state-of-the-art open-source and proprietary methods, followed by qualitative analyses and detailed ablation studies to assess the effectiveness of the proposed reward design.

\subsection{Experimental Setup}
\label{subsec:setup}
\noindent\textbf{Dataset:} 
We conduct experiments on the \textbf{SkeTikZ dataset}~\cite{saito2025sketch2diagram}, which consists of $3,231$ paired samples of hand-drawn sketches and their corresponding TikZ code. The dataset is carefully curated and spans a diverse range of diagram categories, including flowcharts, neural network architectures, and block diagrams, thereby offering a challenging benchmark for sketch-to-code generation. We follow the standard training, validation and test splits provided in~\cite{saito2025sketch2diagram} to ensure fair and consistent comparisons with prior work. To further assess \emph{generalisation}, we additionally evaluate our model on the \textbf{ FlowChart dataset}~\cite{flowchart_db} test set. Notably, this dataset does not provide corresponding ground-truth code, making it suitable for evaluating the robustness of generated outputs in a setting where direct supervision is unavailable. Finally, we note that the training data and methodologies of proprietary GPT-based closed models are not publicly disclosed; therefore, it is difficult to rule out potential overlap between training and testing data when comparing against such systems.

\vspace{1mm}
\noindent\textbf{Baseline Models:} 
We validate our approach through comparisons with a diverse set of baseline models. Specifically, we evaluate against specialised methods such as DeTikZify~\cite{belouadi2024detikzify} and Sketch2Diagram~\cite{saito2025sketch2diagram}, the latter being a strong prior state-of-the-art model tailored for visual TikZ generation. To benchmark performance against the upper bounds of current vision-language models, we additionally compare with state-of-the-art proprietary systems, including GPT-4o-mini, GPT-4o, and GPT-5, using carefully designed zero-shot prompting strategies for diagram generation. Furthermore, we conduct detailed ablation studies of our framework to isolate and quantify the contribution of the proposed multi-reward reinforcement learning components.

\vspace{1mm}
\noindent\textbf{Training and Inference:}
We initialise our experiments using the pre-trained model variants from DeTikZify~\cite{belouadi2024detikzify}, which we further adapt through supervised fine-tuning (SFT) on hand-drawn sketch inputs. The resulting SFT model serves as the base policy for subsequent reinforcement learning (RL). Building on this, we explore multiple reward configurations to assess the impact of our proposed multi-reward framework during RL training. We employ a batch size of $16$ for both the SFT and RL stages, and conduct all experiments on a single NVIDIA A100 GPU for both training and inference. Additionally, following ImgTikZ~\cite{saito2025sketch2diagram}, we adopt a multi-candidate generation strategy with the number of generated candidates set to $10$, enabling more robust selection during evaluation.

\vspace{1mm}
\noindent\textbf{Evaluation Metrics:} 
Given the multi-modal nature of the sketch-to-TikZ generation task, evaluation requires jointly assessing both the structural correctness of the generated code and the visual fidelity of the rendered output. Following established benchmarks~\cite{saito2025sketch2diagram}, we employ a set of quantitative metrics to comprehensively evaluate these complementary aspects:
\begin{itemize}
    \item \textbf{Compilation Success Rate (CSR):} 
    The most fundamental requirement for diagram generation is syntactic validity. CSR measures the proportion of successfully compiled TikZ programs relative to the total number of generated TikZ code outputs, using standard \texttt{pdflatex}~\cite{thanh2000pdftex} and \texttt{LuaTeX}~\cite{luatex2024manual} engines without encountering fatal errors. Both engines are applied sequentially until successful compilation.
    \item \textbf{Image Similarity (ImageSim):} For successfully compiled diagrams, we assess perceptual alignment with the ground-truth reference image. This is measured using \textit{ImageSim}, computed from deep feature embeddings, to capture high-level structural similarity and ensure that the geometric layout and spatial relationships of the input sketch are faithfully preserved in the rendered output~\cite{saito2025sketch2diagram}.
    \item \textbf{Code Similarity (CodeSim):} To evaluate syntactic similarity independent of rendering, we compute \textit{CodeSim} as the cosine similarity between the generated code $Y$ and the reference code $Y_r$ in a $256$-dimensional embedding space. These embeddings are obtained using a pre-trained CodeT5 model~\cite{wang2021codet5}, enabling comparison at a semantic and structural level beyond exact token matching.
    \item \textbf{Character Similarity (CharSim):} To evaluate the model’s OCR capabilities and textual grounding, we compute \textit{CharSim} which measures the accuracy of transcribed textual elements (e.g., mathematical expressions and node labels). This metric captures string-level similarity by computing the ROUGE-1 score between text extracted from the rendered diagram and that from the input sketch~\cite{saito2025sketch2diagram}, thereby assessing both recognition quality and alignment with the source content.
\end{itemize}

\subsection{Experimenting with Multi-Reward Ensembles}
We evaluate our multi-modal framework across a range of reward configurations to analyse the impact of objective design and training strategy on policy learning. Our setup includes single-objective baselines, joint multi-objective optimization, and a staged curriculum-based training paradigm:
\begin{itemize}
\item{\textbf{Single-Objective Training}}: To establish baseline behaviour, we train policies using individual reward components in isolation, as described in Section~\ref{sec:mrs}. In each experiment, the policy is optimized exclusively with respect to a single reward throughout the reinforcement learning phase. This setup enables us to analyse the inductive biases introduced by each proxy objective and understand their isolated impact on the learning dynamics.

\item{\textbf{Multi-Objective Training}}: We next consider a joint optimization setting in which all reward components are applied simultaneously from the outset of training. The overall reward, $\mathcal{R}_{total}(x, y)$, is defined as a weighted combination of specialised objective functions, each addressing a distinct modality gap including textual accuracy, syntactic correctness and perceptual alignment. The contribution of each component is regulated through fixed scalar weights to ensure balanced optimization. The choice of hyperparameters are described in supplementary.

\begin{equation}
    \mathcal{R}_{total}(x, y) = \lambda_{cs}\mathcal{R}_{cs} + \lambda_{ocr}\mathcal{R}_{ocr} + \lambda_{cd}\mathcal{R}_{cd} + \lambda_{ss}\mathcal{R}_{ss}
\end{equation}

\item{\textbf{Curriculum-Based Training}}: To explore staged optimization, we design a curriculum learning framework that decomposes training into sequential phases, progressing from coarse to fine-grained optimization. In the initial phase, the policy is trained using a subset of reward functions until convergence, providing a stable foundation for learning key structural priors. In the subsequent phase, training is resumed with finer, more specialised reward signals, while omitting the coarse-stage rewards allowing the model to focus on refining higher-level fidelity and correctness.

\end{itemize}

\begin{table}[t]
    \centering
    \caption{Performance comparison of our proposed Sketch2TikZ model against SOTA baselines on the SkeTikZ~\cite{saito2025sketch2diagram} and FlowChart-3b~\cite{flowchart_db} datasets. Scores in \textbf{bold} indicate the best-performing open-source model, while underlined values denote the overall best performance across all methods. Here, TL and DS correspond to the TinyLlama-1.1B and DeepSeek-Coder-1.3B variants, respectively.
    }
    \label{tab:main_comparison}
    \scalebox{0.73}{
    \begin{tabular}{|lc|cccc|ccc|}
        \hline
        \multirow{2}{*}{\textbf{Model}} & \multirow{2}{*}{\textbf{Params}} & \multicolumn{4}{|c|}{\textbf{SkeTikZ}~\cite{saito2025sketch2diagram}} & \multicolumn{3}{|c|}{\textbf{FlowChart-3b}~\cite{flowchart_db}} \\ \cline{3-9}
          & & CSR$\uparrow$ & ImageSim $\uparrow$ & CodeSim$\uparrow$ & CharSim$\uparrow$ & CSR$\uparrow$ & ImageSim $\uparrow$ & CharSim$\uparrow$ \\ \hline
        \textit{Closed Models} & & & & & & & &\\ 
        GPT-5 & - & 0.9947 & 0.8461 & 0.7592 & \underline{0.7283} & 0.928 & \underline{0.8412} & \underline{0.6965} \\  
        GPT-4o & - & 0.9659 & \underline{0.8539} & 0.7671 & 0.7131 & 0.885 & 0.8229 & 0.6736 \\
        GPT-4o-mini & - & 0.9449 & 0.8152 & 0.7481 & 0.6998 & 0.871 & 0.8186 & 0.6749 \\ \hline   
        \textit{Open Models} & & & & & & & &\\ 
        LLaVA-Next & 8B & 0.6566 & 0.3784 & 0.6741 & 0.2453 & 0.508 & 0.4355 & 0.2512\\
        Qwen-3 & 4B & 0.9375 & 0.8164 & 0.7670 & 0.6325 & 0.853 & 0.7864 & 0.5871\\
        DeTikZify-TL & 1.1B & 0.9393 & 0.7868 & 0.7602 & 0.5066 & 0.865 & 0.7456 & 0.5733\\
        DeTikZify-DS & 1.3B & 0.9474 & 0.7893 & 0.7798 & 0.5297 & 0.875 & 0.7599 & 0.5846 \\
        \textbf{Ours-TL (SFT)} & 1.1B & 0.969 & 0.8027 & 0.7894 & 0.5638 & 0.905 & 0.7635 & 0.6092 \\ 
       \textbf{Ours-DS (SFT)} & 1.3B & 0.9724 & 0.8139 & \underline{\textbf{0.7966}} & 0.5883 & 0.928 & 0.7889 & 0.6043 \\ 
        \textbf{Ours-TL (RL)} & 1.1B & 0.9975 & 0.8324 & 0.7495 & 0.6302 & 0.936 & 0.8066 & 0.6247 \\
        \textbf{Ours-DS (RL)} & 1.3B & \underline{\textbf{0.9988}} & \textbf{0.8448} & 0.7598 & \textbf{0.6487} & \underline{\textbf{0.945}} & \textbf{0.8205} & \textbf{0.6349} \\ \hline
    \end{tabular}}
\end{table}


\begin{table}[htbp]
\centering
\caption{Qualitative results on the SkeTikZ test set comparing sketch-to-code conversion across different models. The rendered outputs from the generated TikZ code are shown for visual comparison. Both GPT-4o model and DeTikZify fails in case of sketch inputs.}
\begin{tabular}{cccccc}
\hline
Input & GPT-5 & GPT-4o & DeTikZify & Ours (SFT) & Ours (RL) \\
\hline

\includegraphics[width=1.75cm,height=1.75cm]{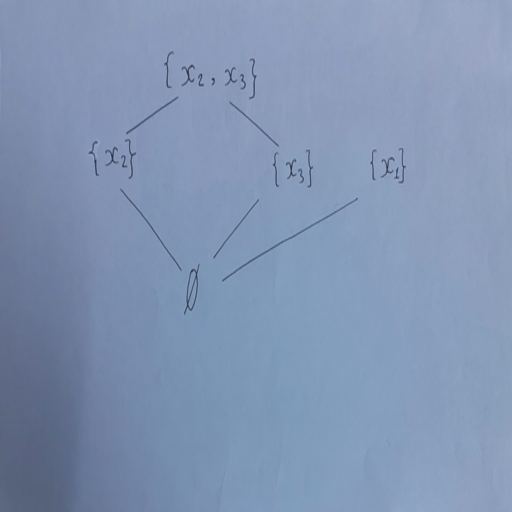} &
\includegraphics[width=1.75cm,height=1.75cm]{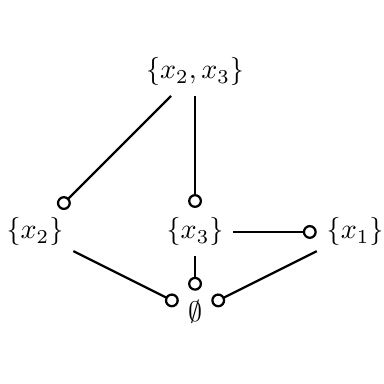} &
\includegraphics[width=1.75cm,height=1.75cm]{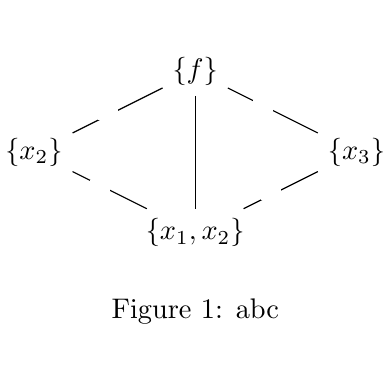} &
\includegraphics[width=1.75cm,height=1.75cm]{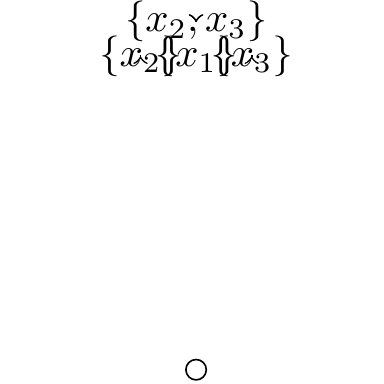} &
\includegraphics[width=1.75cm,height=1.75cm]{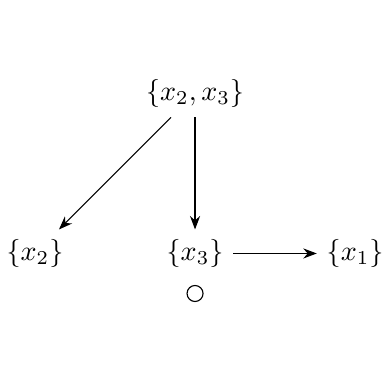} &
\includegraphics[width=1.75cm,height=1.75cm]{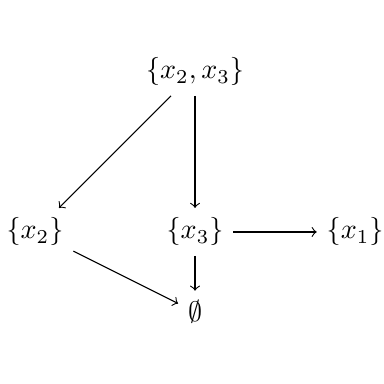} \\

\includegraphics[width=1.75cm,height=1.75cm]{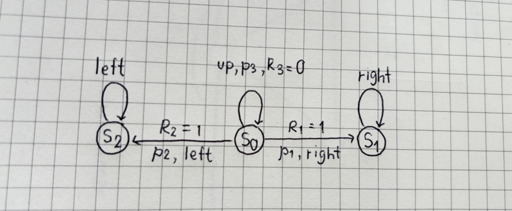} &
\includegraphics[width=1.75cm,height=1.75cm]{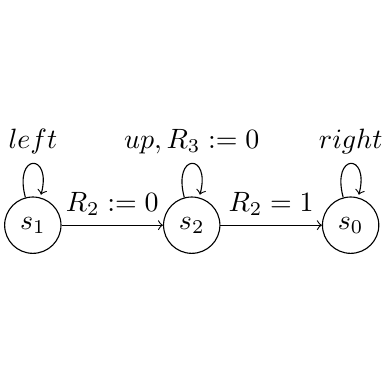} &
\includegraphics[width=1.75cm,height=1.75cm]{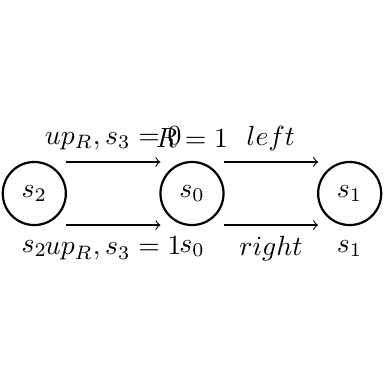} &
\includegraphics[width=1.75cm,height=1.75cm]{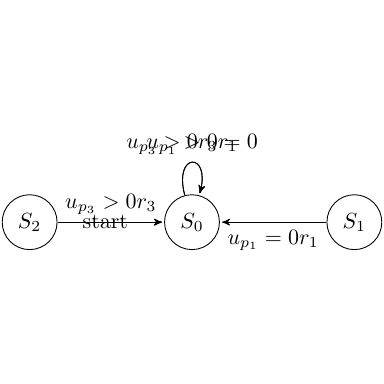} &
\includegraphics[width=1.75cm,height=1.75cm]{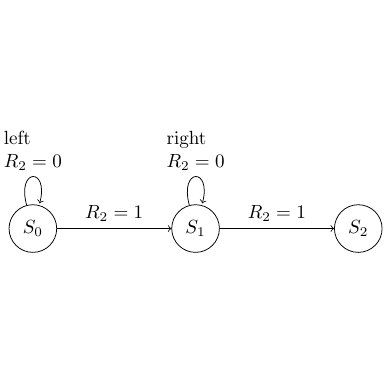} &
\includegraphics[width=1.75cm,height=1.75cm]{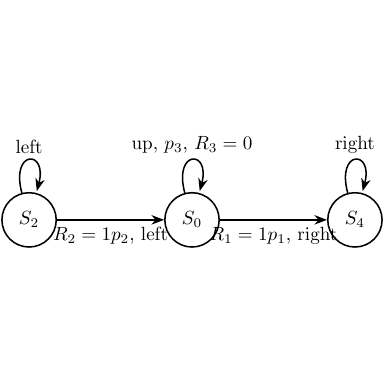} \\

\includegraphics[width=1.75cm,height=1.75cm]{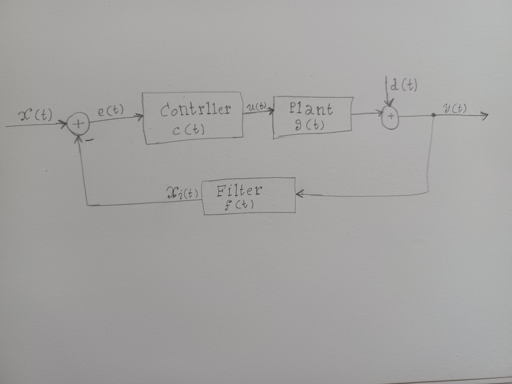} &
\includegraphics[width=1.75cm,height=1.75cm]{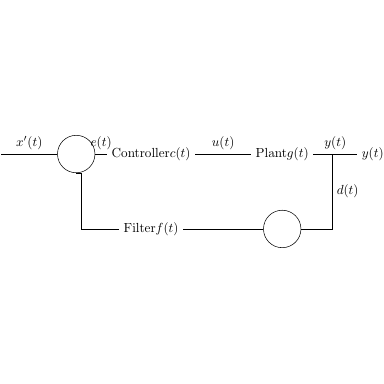} &
\includegraphics[width=1.75cm,height=1.75cm]{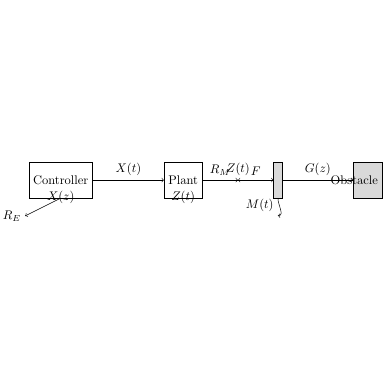} &
\includegraphics[width=1.75cm,height=1.75cm]{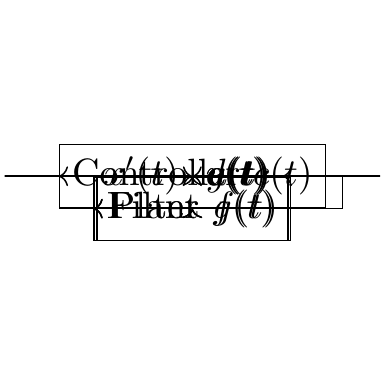} &
\includegraphics[width=1.75cm,height=1.75cm]{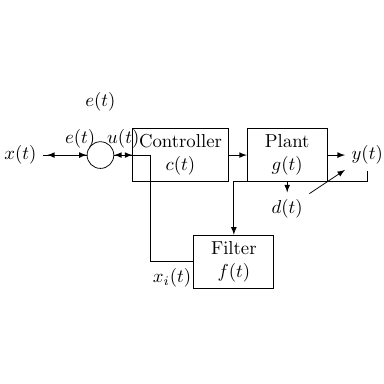} &
\includegraphics[width=1.75cm,height=1.75cm]{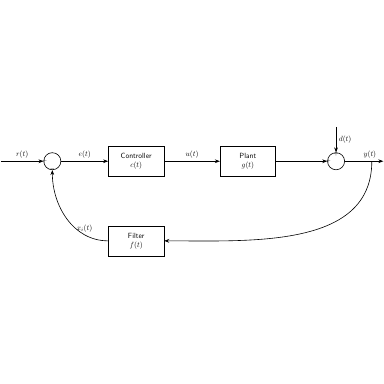} \\


\hline
\end{tabular}
\label{tab:qualitative-results}
\end{table}

\begin{table}[t]
    \centering
    \caption{Ablation study evaluating different reward configurations within the RL framework, including single-objective, multi-objective, and curriculum-based settings. Best-performing scores are highlighted in \textbf{bold}, while second-best results are underlined.}
    \label{tab:ablation}
    \scalebox{0.85}{
    \begin{tabular}{|p{3.0cm}ccccp{2.75cm}|}
    \hline
    \textbf{Training Strategy} & \textbf{CSR$\uparrow$} & \textbf{ImageSim $\uparrow$} & \textbf{CodeSim$\uparrow$} & \textbf{CharSim$\uparrow$}  & \textbf{Convergence Steps ($\times 10^3$) $\downarrow$}  \\ \hline
        \multicolumn{6}{|c|}{\textit{Single-Objective Training}} \\
        $\mathcal{R}_{cs}$ & 0.9978 & 0.7746 & 0.7283 & 0.4771 & 1.25 \\
        $\mathcal{R}_{ocr}$ & 0.9722 & 0.8093 & 0.7642 & 0.5931 & 2.47 \\
        $\mathcal{R}_{ss}$ & 0.9652 & 0.8226 & 0.7549 & 0.5815 & 1.54 \\
        $\mathcal{R}_{cd}$ & 0.9671 & 0.8289 & 0.7661 & 0.5735 & 1.86 \\ \hline
        \multicolumn{6}{|c|}{\textit{Multi-Objective Training}} \\
        $\mathcal{R}_{cs}$+$\mathcal{R}_{ocr}$+$\mathcal{R}_{ss}$+$\mathcal{R}_{cd}$ & 0.9812 & 0.8372 & 0.7436 & 0.6197 & 3.57 \\ \hline
        \multicolumn{6}{|c|}{\textit{Curriculum-Based Training}} \\
        $\mathcal{R}_{cs}$$+$$\mathcal{R}_{ss}$$\rightarrow$$\mathcal{R}_{ocr}$ & 0.9873 & 0.8219 & 0.7355 & 0.6257  & 3.15  \footnotesize(1.45+1.7) \\
        $\mathcal{R}_{cs}$$+$$\mathcal{R}_{cd}$$\rightarrow$$\mathcal{R}_{ocr}$ & 0.9938 & 0.8354 & 0.7488 & 0.6347 & 3.46  \footnotesize(1.63+1.83) \\        
        $\mathcal{R}_{cs}$$+$$\mathcal{R}_{ss}$$\rightarrow$$\mathcal{R}_{cd}$ & \textbf{0.9988} & \textbf{0.8448} & 0.7598 & \textbf{0.6487} & 3.03 \footnotesize (1.45+1.58) \\
        $\mathcal{R}_{cs}$$+$$\mathcal{R}_{cd}$$\rightarrow$$\mathcal{R}_{ss}$$\rightarrow$$\mathcal{R}_{ocr}$ & 0.9975 & 0.8377 & \underline{0.7603} & 0.6396 & 4.31 \footnotesize (1.63+1.11+1.57)\\
        $\mathcal{R}_{cs}$$+$$\mathcal{R}_{ss}$$\rightarrow$$\mathcal{R}_{cd}$$\rightarrow$$\mathcal{R}_{ocr}$ & \underline{0.9982} & \underline{0.8441} & \textbf{0.7665} & \underline{0.6483} & {4.67 \footnotesize (1.35+1.58+1.74)}\\
        \hline
    \end{tabular}}
\end{table}

\section{Results and Discussions}
\subsection{Performance Comparison with SOTA methods}
We first present a quantitative comparison of our trained model against state-of-the-art methods on the SkeTikZ test set~\cite{saito2025sketch2diagram}. To ensure fairness, we adopt comparable fine-tuned model variants across all baselines. As the model weights of ImgTikZ~\cite{saito2025sketch2diagram} are not publicly available, we instead initialise from the publicly released DeTikZify~\cite{belouadi2024detikzify} checkpoints and follow a similar training procedure, adapting smaller model variants to align with the scale and capacity of the ImgTikZ framework. As shown in Table~\ref{tab:main_comparison}, SFT models demonstrate consistent improvements across all metrics compared to their corresponding DeTikZify variants (1.1B / 1.3B), achieving gains of 2.97\% / 2.5\% in $CSR$, 1.59\% / 2.46\% in ImageSim, 2.92\% / 1.68\% in CodeSim, and 5.72\% / 5.86\% in CharSim. Further, among the evaluated RL configurations, the best performance is obtained using the curriculum-based training strategy. This model improves over the SFT baseline with improvements of 2.85\% / 2.64\% in $CSR$, 2.97\% / 3.09\% in ImageSim, and 6.64\% / 6.04\% in CharSim, although a slight decrease is observed in CodeSim (-4.0\% / -3.68\%). We also compared our method with LlaVa-Next and Qwen model.
When compared against proprietary GPT-based models, our approach yields notably higher performance, improving $CSR$ by 3.29\% / 5.39\% (GPT-4o / GPT-4o-mini), ImageSim by 2.96\% (GPT-4o-mini), and CodeSim by 1.17\% (GPT-4o-mini).We also provide a qualitative comparison of diagrams rendered from the generated TikZ code for our method against baseline and proprietary models, as shown in Table~\ref{tab:qualitative-results}. The results demonstrate that our approach more effectively converts hand-drawn sketches compared to prior methods. Notably, DeTikZify fails to generalize to sketch inputs as it is primarily trained on rendered images rather than hand-drawn sketches.


To further evaluate the robustness of our model, we conduct zero-shot evaluation over $70$ examples from FlowChart-3b dataset~\cite{flowchart_db}, which primarily consists of flowcharts.  The results show that our best-performing comparable model (in terms of parameters) outperforms existing state-of-the-art methods by 7.0\% ($CSR$), 6.06\% (ImageSim) and 5.03\% (CharSim). In comparison with proprietary GPT-based models, our approach achieves competitive performance, with notable improvements 
in $CSR$ by 1.7\% / 6.0\% / 7.4\% (GPT-5/ GPT-4o/ GPT-4o-mini) and marginal gain of 0.19\% in ImageSim over GPT-4o-mini.

\subsection{Performance Analysis Across Reward Configurations}
\noindent\textbf{Single Objective Optimization}: We first evaluate single-objective training to isolate the influence of each proxy reward, and the results, as shown in Table~\ref{tab:ablation} clearly confirm their expected limitations. optimizing solely with the code-structure reward $\mathcal{R}_{cs}$ yields near-perfect compilability ($CSR=0.9978$) and the fastest convergence ($1.25\times10^3$ steps), but leads to substantially weaker perceptual and semantic alignment (ImageSim $=0.7746$, CodeSim $=0.7283$, CharSim $=0.4771$). This indicates that structurally valid outputs do not necessarily reflect the visual or textual intent of the input sketch. In contrast, training with only perceptual objectives improves visual similarity but compromises structural reliability and optimization efficiency. For example, the OCR reward improves ImageSim ($0.8093$) and CodeSim ($0.7642$), but reduces $CSR$ to $0.9722$ and nearly doubles the convergence time ($2.47\times10^3$ steps). Similarly, the similarity- and diffusion-based rewards further enhance perceptual alignment (ImageSim up to $0.8226$ and $0.8289$, respectively), yet fail to maintain structural robustness ($CSR=0.9652$ and $0.9671$) and require more training steps to converge ($1.54\times10^3$ and $1.86\times10^3$). Overall, these findings demonstrate that optimizing individual objectives leads to imbalanced behaviour: structure-only training sacrifices semantic fidelity, whereas perceptual rewards improve appearance at the cost of compilability and robustness.\\

\noindent\textbf{Multi-Objective Optimization}: As evident in Table~\ref{tab:ablation}, this approach ($\mathcal{R}_{cs} + \mathcal{R}_{ocr} + \mathcal{R}_{ss} + \mathcal{R}_{cd}$) improves perceptual fidelity over all single-objective configurations, achieving the highest ImageSim (0.8372), surpassing the best individual result (0.8289 with $\mathcal{R}_{cd}$), and also increasing CharSim to 0.6197 compared to the single-objective range (0.4771 to 0.5931). However, these improvements come at the cost of structural reliability and optimization efficiency. Specifically, $CSR$ decreases to 0.9812, notably lower than the structure-only optimum (0.9978), while convergence slows significantly to $3.57\times10^3$ steps compared to the fastest single-objective setting ($1.25\times10^3$, $\mathcal{R}_{cs}$). Furthermore, CodeSim does not show consistent gains, reaching 0.7436, which remains below the performance of both $\mathcal{R}_{cd}$ (0.7661) and $\mathcal{R}_{ocr}$ (0.7642).
These results highlight the inherent trade-off in simultaneous multi-objective optimization, where improvements in perceptual alignment may lead to diminished structural robustness and increased training complexity.

These findings empirically highlight the presence of optimization interference. While the joint objective mitigates the extreme failure modes observed in single-reward training, it introduces a trade-off where competing gradient signals hinder the simultaneous maximization of all metrics. In particular, the drop in $CSR$, coupled with increased convergence time, suggests that structural learning is partially compromised when co-optimized alongside high-frequency perceptual rewards. This effect aligns with the observed limitations in fine-grained spatial alignment, indicating that although simultaneous optimization yields more balanced outcomes, it remains susceptible to gradient conflicts that ultimately constrain overall performance.\\

\noindent\textbf{Curriculum-based Optimization}:
Table~\ref{tab:ablation} demonstrates that these staged strategies consistently improve both structural reliability and perceptual fidelity compared to multi-objective optimization ($CSR$=0.9812, ImageSim=0.8372). The two-phase curriculum setting ($\mathcal{R}_{cs}+\mathcal{R}_{ss}\rightarrow \mathcal{R}_{cd}$) achieves near-perfect compilability (0.9988) while simultaneously improving perceptual and structural alignment, with higher ImageSim (0.8448) and CodeSim (0.7598), outperforming the joint optimisation strategy across all metrics. Notably, these gains are achieved with lower computational cost, requiring fewer optimization steps ($3.03\times10^3$ compared to $3.57\times10^3$), indicating more efficient convergence. Similar trends are observed across other curriculum variants (e.g., $\mathcal{R}_{cs}+\mathcal{R}_{cd}\rightarrow \mathcal{R}_{ocr}$), achieving $CSR=0.9938$ and ImageSim = 0.8354), further validating that emphasising coarse structural learning before introducing fine-grained rewards leads to more stable and effective training dynamics.\\


\noindent\textbf{Efficacy of Sequential Application}: The curriculum-based training consistently yields the strongest overall trade-offs, effectively dominating both single-objective and multi-objective settings. Notably, deeper curricula such as $\mathcal{R}_{cs}+\mathcal{R}_{ss}\rightarrow \mathcal{R}_{cd} \rightarrow \mathcal{R}_{ocr}$ achieve the best CodeSim while remaining competitive on other metrics ($CSR=0.9982$, ImageSim = 0.8441, CodeSim = 0.7665, CharSim = 0.6483), while best CSR/ImageSim/CharSim belong to $\mathcal{R}_{cs}+\mathcal{R}_{ss}\rightarrow \mathcal{R}_{cd}$ and it is used for comparison in Table~\ref{tab:main_comparison}. Curriculum-based configuration surpasses the best single-objective results (e.g., CodeSim 0.7665 vs. 0.7661 with $\mathcal{R}_{cd}$), demonstrating its ability to simultaneously optimise multiple objectives. Similarly, three-stage schedules like $\mathcal{R}_{cs}+\mathcal{R}_{cd}\rightarrow \mathcal{R}_{ss} \rightarrow \mathcal{R}_{ocr}$ maintain high CSR (0.9975) while improving semantic alignment (CharSim 0.6396). These results highlight the effectiveness of temporally decoupling objectives: the model first learns stable global structure and syntax, and subsequently refines spatial and semantic details without degrading earlier gains. Overall, this staged optimization paradigm clearly outperforms simultaneous multi-objective training by reducing gradient interference and enabling more efficient and balanced learning.

\begin{table}[t]
    \centering
    \caption{Over-optimization analysis comparing internal proxy rewards against LLM-based gold scores.}
    \label{tab:reward_var_compare}
    \scalebox{0.8}{
    \begin{tabular}{|cp{2.5cm}p{2.5cm}p{2.5cm}p{2.75cm}|}
        \hline
        \textbf{Reward} & \textbf{Mean Reward Score (Start $\to$ End)} & \textbf{Image Similarity (Start $\to$ End)} & \textbf{Gold Score (Start $\to$ End)} & \textbf{Over-optimization Gap ($\Delta$)} \\ \hline 
        $\mathcal{R}_{cs}$ & 0.86 $\to$ 0.96 & 0.8139 $\to$ 0.7746 & 2.84 $\to$ 2.56 & -0.0393 / -0.28 \\
        $\mathcal{R}_{ocr}$ & 0.47 $\to$ 0.51 & 0.8139 $\to$ 0.8093 & 2.84 $\to$ 2.77 & -0.0046 / -0.07 \\  
        $\mathcal{R}_{ss}$ & 0.59 $\to$ 0.66 & 0.8139 $\to$ 0.8226 & 2.84 $\to$ 2.85 & 0.0087 / 0.01 \\
        $\mathcal{R}_{cd}$ & 0.63 $\to$ 0.71 & 0.8139 $\to$ 0.8289 & 2.84 $\to$ 2.81 & 0.015 / -0.03 \\
        $\mathcal{R}_{cs}$+$\mathcal{R}_{ss}$ & 0.72 $\to$ 0.78 & 0.8139 $\to$ 0.8295 & 2.84 $\to$ 2.93 & 0.0156 / 0.09 \\ 
        $\mathcal{R}_{cs}$$+$$\mathcal{R}_{ss}$$\rightarrow$$\mathcal{R}_{cd}$ & 0.73 $\to$ 0.81 & 0.8295 $\to$ 0.8448 & 2.93 $\to$ 3.15 & 0.0153 / 0.22 \\ 
        $\mathcal{R}_{cs}$$+$$\mathcal{R}_{ss}$$\rightarrow$$\mathcal{R}_{cd}$$\rightarrow$$\mathcal{R}_{ocr}$ & 0.71 $\to$ 0.76 & 0.8448 $\to$ 0.8441 & 3.15 $\to$ 3.15 & -0.007 / 0.00 \\ \hline
    \end{tabular}}
\end{table}


\subsection{Ablation Study Analysis}
To rigorously validate our architectural decisions, we conduct targeted ablation studies focusing on the optimization dynamics of our framework. Specifically, we analyze how our multi-reward system prevents policy collapse and how the deterministic nature of our specialized rewards stabilizes the Group Relative Policy Optimization (GRPO) training process.\\

\noindent\textbf{Mitigating Over-optimization via Multi-Reward Ensembles}:
In proxy-based reinforcement learning, policies tend to be highly sensitive to the chosen optimization objective. Aggressively optimizing an imperfect proxy reward can lead the model to exploit spurious correlations, causing a divergence from the true underlying objective. This phenomenon is commonly referred to as \textit{reward over-optimization}~\cite{coste2024rewardmodelensembleshelp}.

In the context of multi-modal Sketch2TikZ generation, single-objective proxy rewards are particularly susceptible to adversarial optimization by the policy network. To systematically characterise this behaviour, we conduct a detailed over-optimization analysis (Table~\ref{tab:reward_var_compare}). The proxy score is a mathematical objective that the policy directly maximizes during RL optimization. We track the proxy score against the LLM-based Gold Score at both the beginning and end of RL fine-tuning. For the LLM-based Gold Reward, we employ a large language model as an automated evaluator, calibrated to approximate human preference. Given the input sketch (reference) and the corresponding rendered TikZ output (candidate), the evaluator assesses (i) semantic fidelity  for capturing the correctness of elements and relationships, and (ii) spatial accuracy for reflecting the alignment of layout and relative positioning. The model assigns a scalar score on a discrete scale from 0 to 5, where higher values indicate closer agreement with the reference. 
We have defined here, the over-optimization gap ($\Delta$) as the change in ImageSim and the change in Gold Score (Start$\rightarrow$End) as the reward converges. Table~\ref{tab:reward_var_compare} demonstrates different experiment settings where mean reward score shows the initial reward before training and rewards after convergence. Similarly Image similarity scores and gold scores are computed for the same model states. Positive over-optimization gap shows the rewards is helping in improving model, while negative or close to zero values, demonstrates detrimental or no learning. 

\section{Conclusion}
We presented Sketch2TikZ, a reinforcement learning framework for converting hand-drawn sketches into executable TikZ programs. Our study demonstrates that sketch-to-code generation cannot be effectively optimized through token-level supervision or a single proxy objective, as valid outputs must simultaneously satisfy multiple constraints, including compilability, visual fidelity, textual correctness and structural consistency. To address this, we introduced a sequential multi-reward GRPO training strategy that combines rewards such as Code-Structure, OCR, CleanDIFT, and Self-Similarity enabling the model to receive targeted feedback. Importantly, these rewards are designed as continuous soft feedback signals with strict hard constraints, allowing the policy to improve smoothly while avoiding optimization cliffs and preventing invalid solutions from being rewarded. Across SkeTikZ and FlowChart-3b benchmarks, Sketch2TikZ achieves near-perfect compilation success while improving visual and textual alignment over strong open-source baselines and remaining competitive with proprietary vision-language models. Our ablations further show that curriculum-based reward scheduling outperforms both single-objective and simultaneous multi-objective training by reducing reward interference, improving convergence stability and mitigating proxy over-optimization. These results demonstrate that carefully designed, reference-aware reward signals can substantially improve structured diagram generation from noisy sketches. Going forward, we will focus on scaling the framework to larger models and datasets to further improve generalization and compositional reasoning.

\bibliography{egbib}
\end{document}


\maketitle

\appendix
\section{Implementation details}
\subsection{Tools} 
We used both PdfLaTeX \textbf{(pdfTeX 3.141592653-2.6-1.40.25 (TeX Live 2023/Debian))} and LuaLaTeX \textbf{(LuaHBTeX, Version 1.17.0 (TeX Live 2023/Debian) Dev id: 7581)} to compile the generated TikZ code. The compilation process first attempts PdfLaTeX, and if unsuccessful, falls back to LuaLaTeX, using whichever produces a valid output.

Unwanted line numbers which appear in the rendered output, are removed by explicitly disabling page styling in generated TikZ code. 
Then we use python package \textit{pdfCropMargins}, to crop the margins of PDF files, which is then converted to an image file, for further steps.

\subsection{Hyperparameters for inference} 
Across all evaluation experiments, input sketch is an RGB image with $384 \times 384$ resolution. We have configured the language model generation process with $max\_new\_tokens = 2048$. Sampling was enabled, using nucleus sampling with $top_p = 0.95$, no top-$k$ restriction ($top_k = 0$), and a temperature of $0.9$.

\begin{figure*}[ht]
    \centering
    \includegraphics[width=\linewidth]{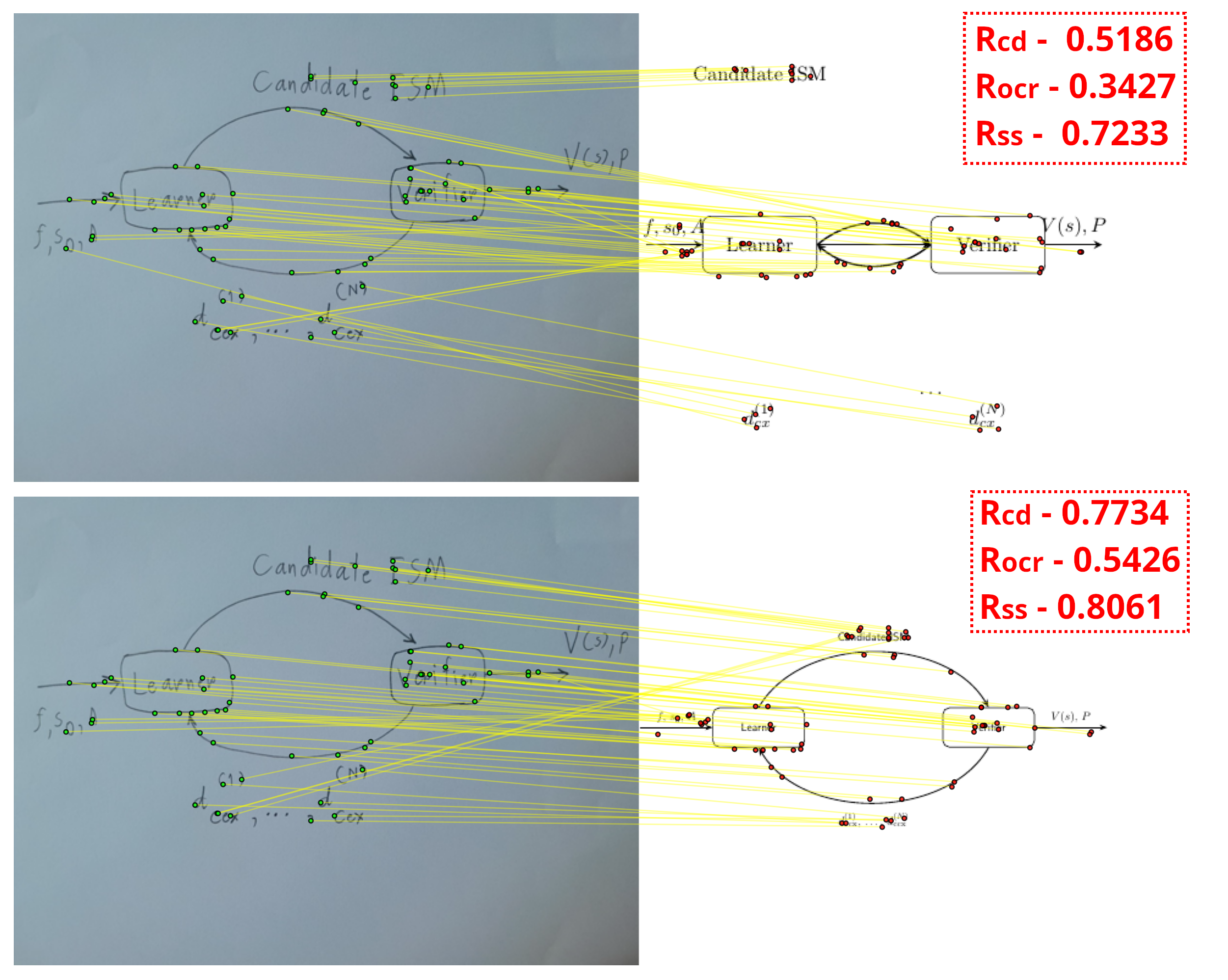}
    \caption{Figure shows two scenarios (rows), where left is the input sketch image, and right is the rendered image from code. Yellow lines show the correspondence established between edge points from sketch to rendered images, using CleanDIFT. Corresponding reward values obtained are depicted in the figure.}
    \label{fig:reward_comparison}
\end{figure*}

\subsection{Hyperparameters used in Rewards}
\subsubsection{Code-Structure reward}
We have experimented with multiple values of $\alpha$ and $r_{base}$. The optimal value used for experiment is $\alpha = 0.1$ and $r_{base} = 2$.

\subsubsection{OCR reward}
\paragraph{OCR Engine}
Hunyuan OCR~\cite{hunyuanvisionteam2025hunyuanocrtechnicalreport} was used for text detections in rendered and sketch images. We have used the resolution of $1024 \times 1024$ to scale the image before processing OCR.
\paragraph{Soft-IoU}
We have defined the soft spatial signal to be computed over normalised coordinates. We have defined the value for $\alpha_{ocr} = 20$, which provides soft signal even when the matching texts are nearby and not overlapping.

\begin{figure}[t]
\centering
\setlength{\tabcolsep}{1pt}
\renewcommand{\arraystretch}{0}
\vspace{5pt}
\begin{tabular}{cccccc}
\textbf{Input} & \textbf{GPT-5} & \textbf{GPT-4o} & \textbf{DeTikZify} & \textbf{SFT} & \textbf{SFT-RL}  \\ 

\includegraphics[width=0.16\linewidth]{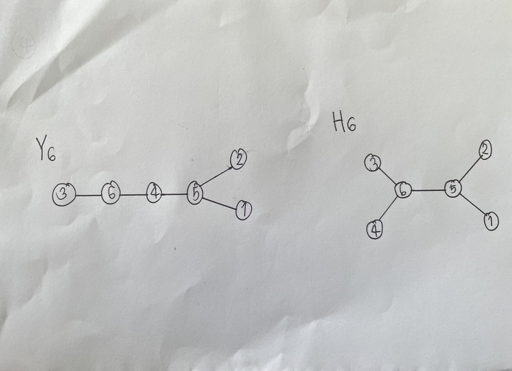} &
\includegraphics[width=0.16\linewidth]{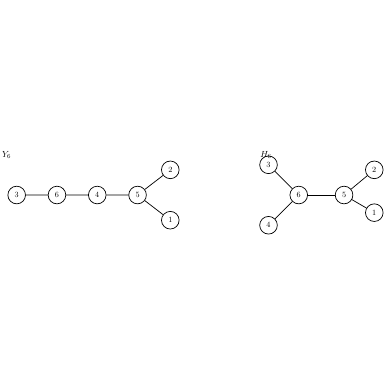} &
\includegraphics[width=0.16\linewidth]{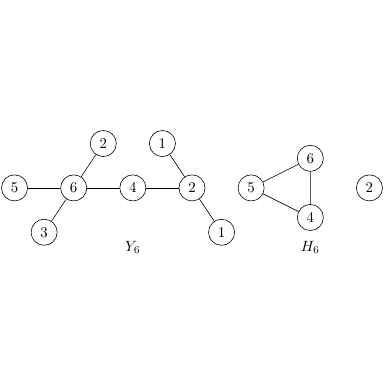} &
\includegraphics[width=0.16\linewidth]{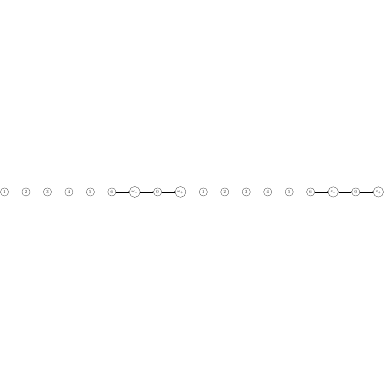} &
\includegraphics[width=0.16\linewidth]{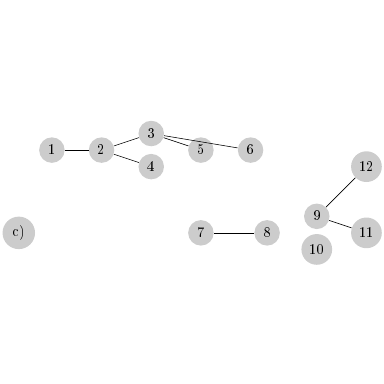} &
\includegraphics[width=0.16\linewidth]{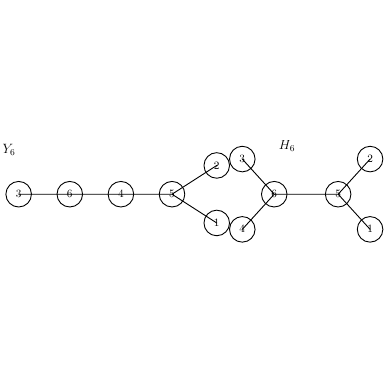} \\ \\

\includegraphics[width=0.16\linewidth]{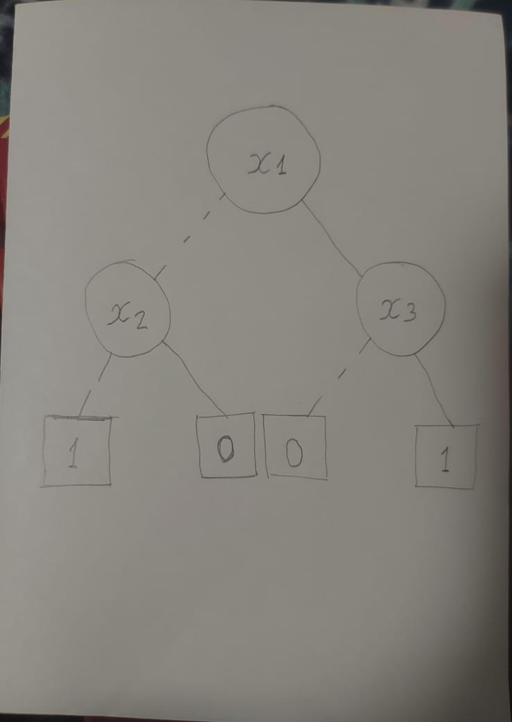} &
\includegraphics[width=0.16\linewidth]{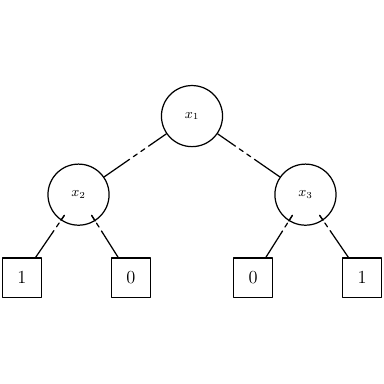} &
\includegraphics[width=0.16\linewidth]{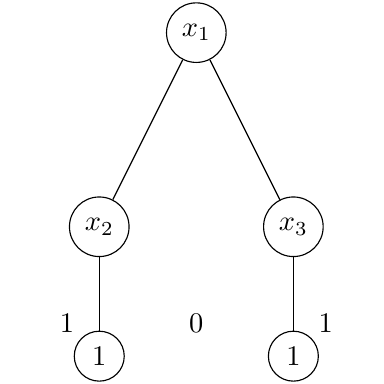} &
\includegraphics[width=0.16\linewidth]{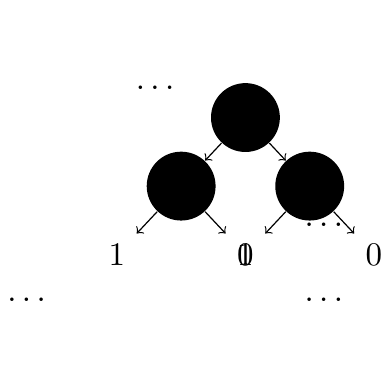} &
\includegraphics[width=0.16\linewidth]{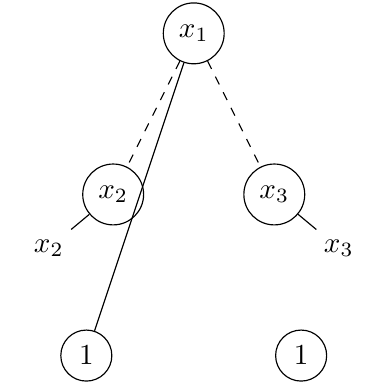} &
\includegraphics[width=0.16\linewidth]{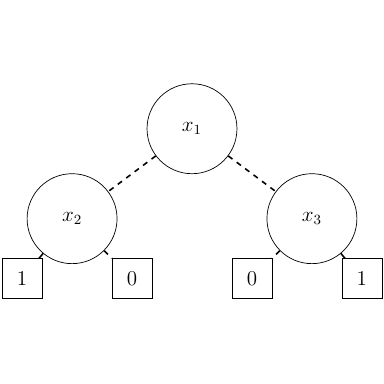} \\ \\ 

\includegraphics[width=0.16\linewidth]{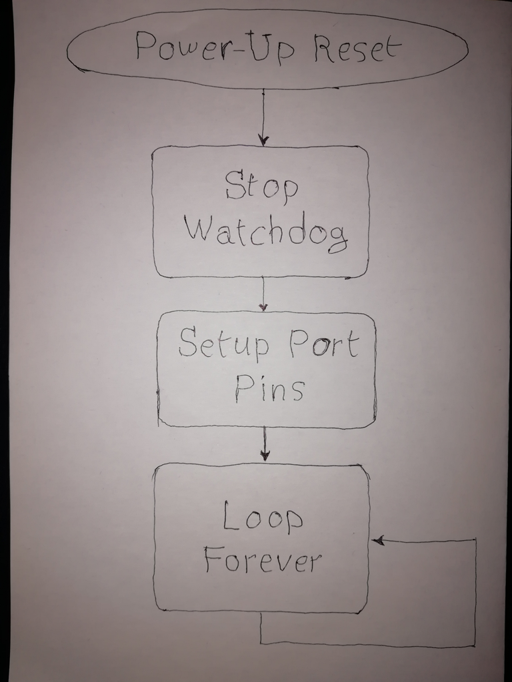} &
\includegraphics[width=0.16\linewidth]{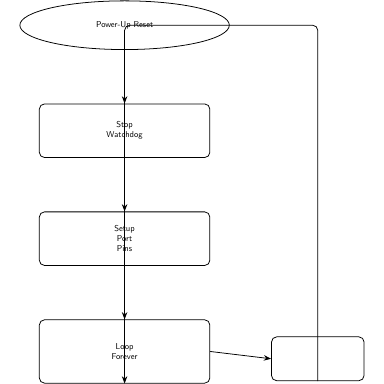} &
\includegraphics[width=0.16\linewidth]{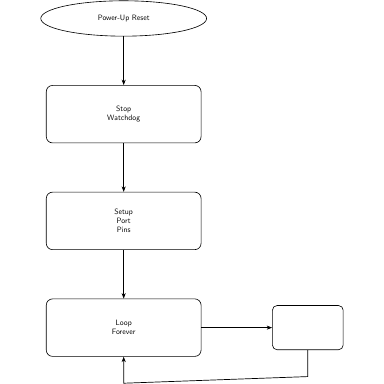} &
\includegraphics[width=0.16\linewidth]{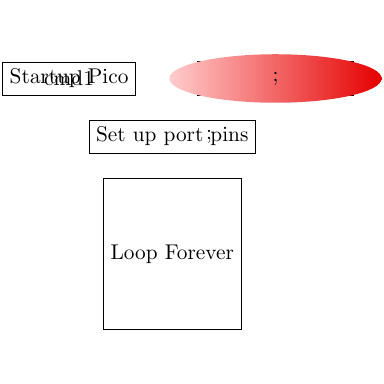} &
\includegraphics[width=0.16\linewidth]{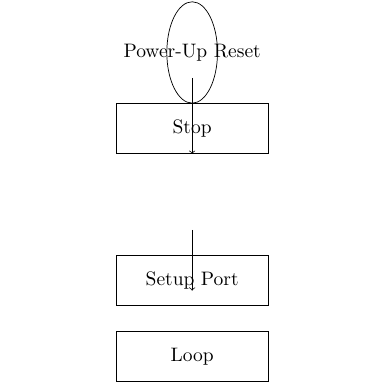} &
\includegraphics[width=0.16\linewidth]{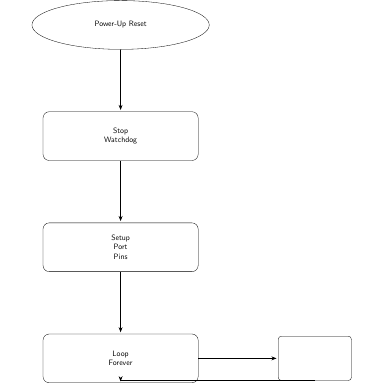} \\ \\

\includegraphics[width=0.16\linewidth]{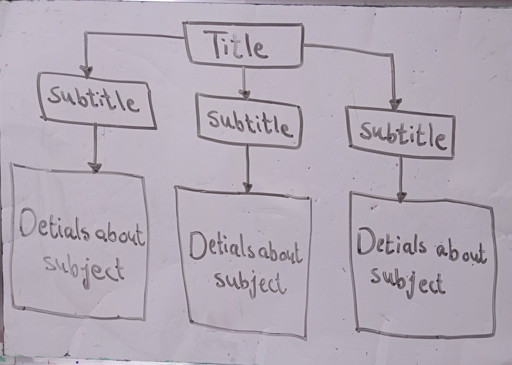} &
\includegraphics[width=0.16\linewidth]{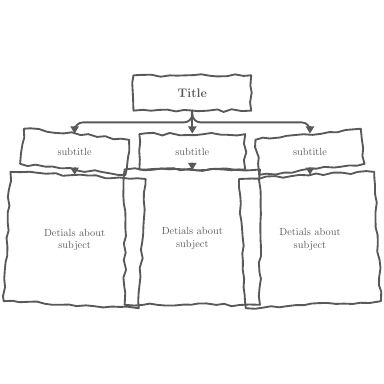} &
\includegraphics[width=0.16\linewidth]{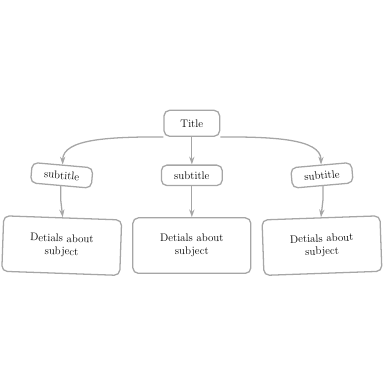} &
\includegraphics[width=0.16\linewidth]{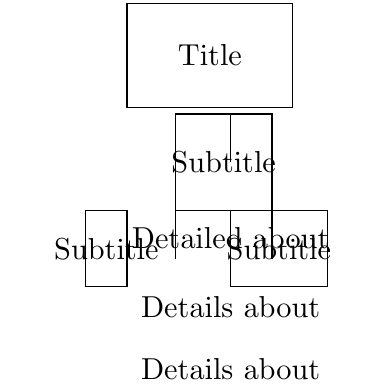} &
\includegraphics[width=0.16\linewidth]{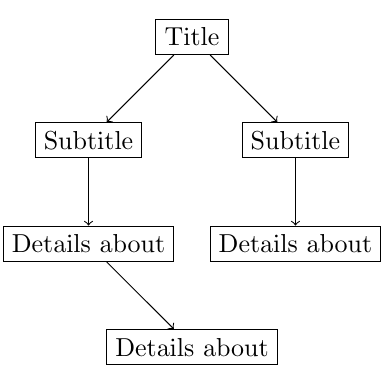} &
\includegraphics[width=0.16\linewidth]{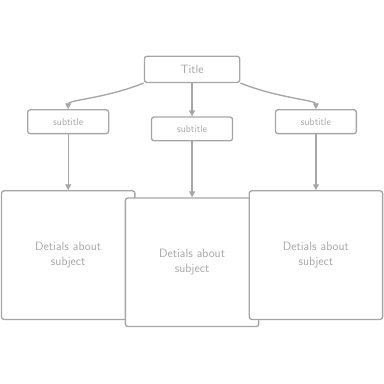} \\

\end{tabular}

\caption{Qualitative comparison across models. Each row corresponds to a different input example. Columns show results from GPT-5, GPT-4o, DeTikZify, SFT, and SFT-RL.}
\label{fig:qual_comparison}
\end{figure}

\subsubsection{CleanDIFT reward}
We sample $M=100$ foreground edge points uniformly from the binarized sketch and match each point to similarly extracted points in the rendered image by nearest-neighbor search in CleanDIFT feature space using cosine similarity. 
\paragraph{Binarizing image}
We thresholded the image using OpenCV library, \textit{cv2.adaptiveThreshold} method with

\begin{lstlisting}[language=Python, basicstyle=\ttfamily\footnotesize]
maxValue = 255
adaptiveMethod = cv2.ADAPTIVE_THRESH_MEAN_C
thresholdType = cv2.THRESH_BINARY
blockSize = 9
C = 50
\end{lstlisting}

\paragraph{Sampling points from Edges}
Points are sampled from the edges (as the sketches are sparse) of the image, such that
\begin{lstlisting}[language=Python, basicstyle=\ttfamily\footnotesize]
edge_pixels = np.argwhere(image == 255)
\end{lstlisting}

\subsubsection{Self Similarity reward}
In the self similarity reward, we apply a $3{\times}3$ max-pooling operator $P$ with stride = $1$. 
For all $m \in M$ models where $M \in [CLIP,DINO, OpenCLIP]$, we use patch tokens from the final transformer block prior to global pooling ($l = L{-}1$), discarding the CLS token.

To visualize the impact of our multiple rewards used in our curriculum-based multi-objective RL, we demonstrate the obtained reward values for two rendered figures which are relatively similar, in Figure~\ref{fig:reward_comparison}.

\subsection{Hyperparameters used in Multi-Reward Ensembles}
For all cases of the Single-Objective Training, we have used fixed scaling factor of $3$ for all the rewards.

While, for multi-objective training and curriculum based training, we have used $\lambda_{cs} = 3.0$ to give priority for compilation, while the rewards for  $\lambda_{cd}$, $\lambda_{ss}$ and $\lambda_{ocr}$ are kept fixed at $1.5$, $1.2$ and $2.0$ respectively. The rewards are chosen by observing mean rewards values over ground truth pairs, and maintaining equal impact of all rewards.

\begin{figure}[t]
    \centering
    \includegraphics[width=0.48\linewidth]{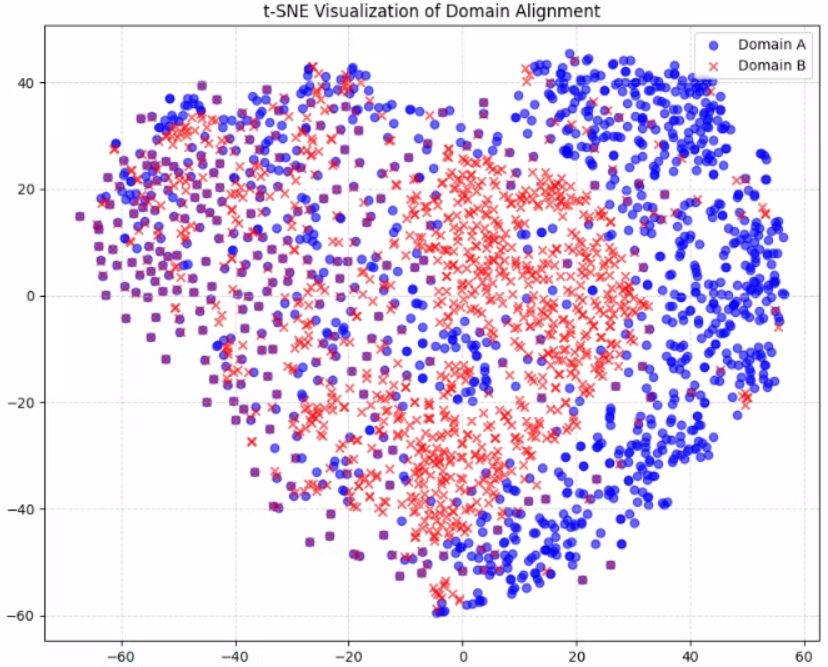}
    \hfill
    \includegraphics[width=0.48\linewidth]{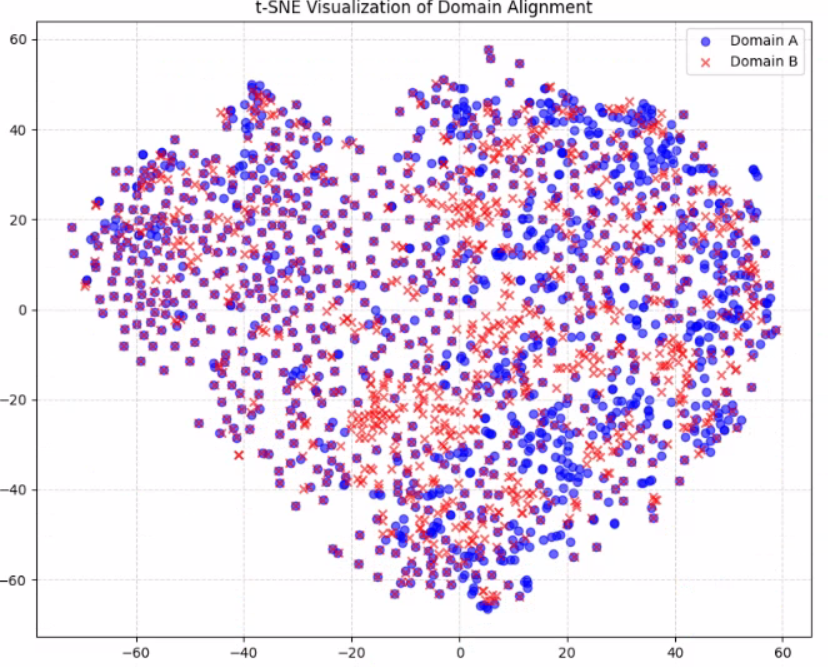}
    \caption{t-SNE plots of multimodal adapter features. The left figure shows feature distributions before training, while the right figure illustrates improved cross-domain alignment after training.}
    \label{fig:testing_adapter}
\end{figure}

\section{Supervised Fine Tuning}
We have used the base model from DeTikZify~\cite{belouadi2024detikzify}, which is fine-tuned over SkeTikZ~\cite{saito2025sketch2diagram}. Since the base model is primarily trained over rendered images, we have to adapt the model to process hand-drawn sketch images.
Therefore, we fine-tune a lightweight multimodal adapter from the original code, using a teacher–student setup on the original 2.5k paired dataset with augmentations. Rest of the model except the adapter layer is kept frozen.

A pre-trained adapter from original model serves as the teacher, while a newly initialized adapter is trained as the student.

\begin{figure}[]
\centering
\setlength{\tabcolsep}{3pt}
\renewcommand{\arraystretch}{0}

\begin{tabular}{ccc}
\textbf{Input} & \textbf{Failure 1} & \textbf{Failure 2} \\

\includegraphics[width=0.3\linewidth, height=2.8cm]{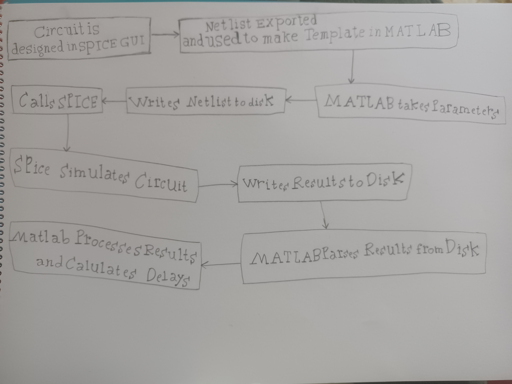} &
\includegraphics[width=0.3\linewidth, height=2.8cm]{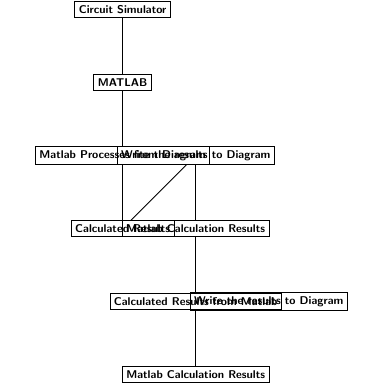} &
\includegraphics[width=0.3\linewidth, height=2.8cm]{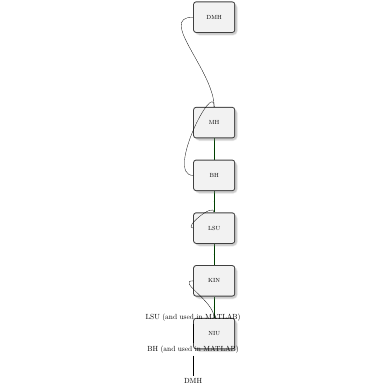} \\

\includegraphics[width=0.3\linewidth, height=2.8cm]{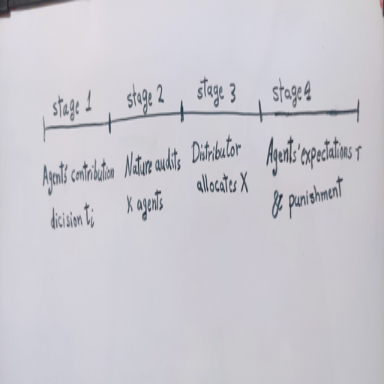} &
\includegraphics[width=0.3\linewidth, height=2.8cm]{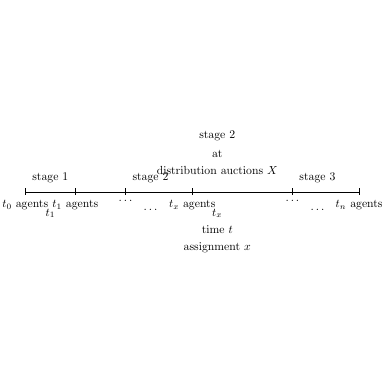} &
\includegraphics[width=0.3\linewidth, height=2.8cm]{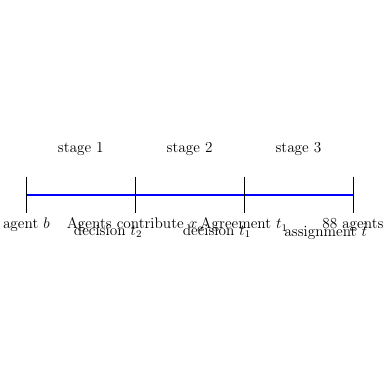} \\

\includegraphics[width=0.3\linewidth, height=2.8cm]{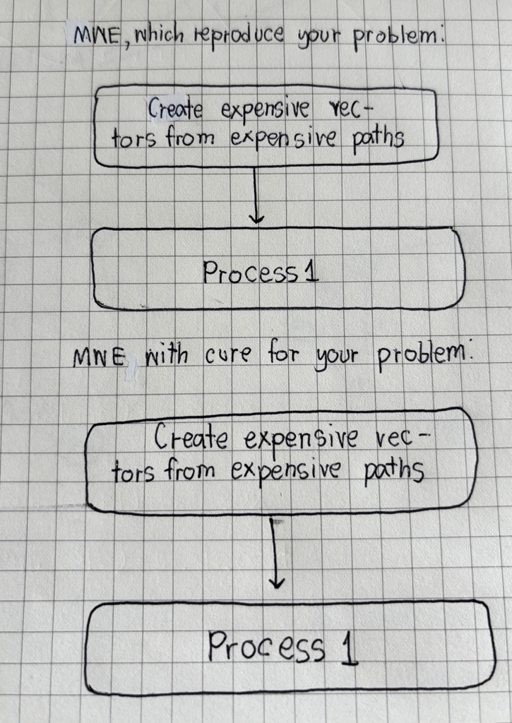} &
\includegraphics[width=0.3\linewidth, height=2.8cm]{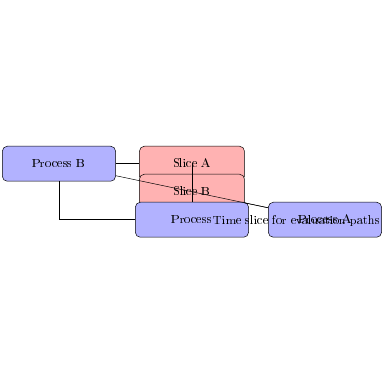} &
\includegraphics[width=0.3\linewidth, height=2.8cm]{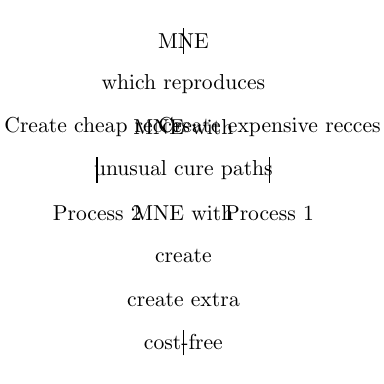} \\

\end{tabular}

\caption{Failure cases. Each row shows an input sketch and two incorrect outputs produced by our method. }
\label{fig:failure_cases}
\end{figure}

Defining formally, let $A$ and $B$ denote two domains representing rendered diagram and hand-drawn sketch diagrams respectively. The student is optimized with two losses. 
The \textbf{preservation loss} enforces consistency with the teacher on domain $A$:
\[
\mathcal{L}_{\text{pres}} = 
\| f_t(A) - f_s(A) \|_2^2,
\]
where $f_t$ and $f_s$ denote teacher and student adapters.

The \textbf{domain alignment loss} encourages the student to produce similar representations across domains:
\[
\mathcal{L}_{\text{align}} = 
1 - \cos\big(f_s(A),\, f_s(B)\big).
\]

The final objective is a weighted combination:
\[
\mathcal{L} = \lambda_p \mathcal{L}_{\text{pres}} + \lambda_a \mathcal{L}_{\text{align}}.
\]
We have used the $\lambda_p=0.5$ and $\lambda_a=0.5$ each, and trained the model for $3$ epochs over $2.5k$ randomly sampled examples from DaTikZ-v2~\cite{belouadi2024detikzify} and SkeTikZ~\cite{saito2025sketch2diagram} dataset respectively. The t‑SNE plot for domain alignment over test set after adapter fine-tuning is shown in Figure~\ref{fig:testing_adapter}.

\paragraph{Data Augmentation.}
We apply a sequence of mild geometric and photometric augmentations during training. We use an additional augmentation of \textit{Color Inversion} apart from using similar transforms from Sketch2Diagram~\cite{saito2025sketch2diagram}.
The transformations are applied with specified probabilities as follows:

\begin{itemize}
    \item \textbf{Random scaling:} anisotropic scaling in the range $[0.7, 1.0]$.
    \item \textbf{Random rotation:} rotation sampled from $[-3^\circ, 3^\circ]$.
    \item \textbf{Random translation:} shifts up to $\pm 5\%$ along both axes.
    \item \textbf{Color inversion (p=0.5):} random inversion of image intensities.
    \item \textbf{Color temperature jitter (p=0.2):} variation in illumination tone.
    \item \textbf{Additive Gaussian noise (p=0.3):} noise with standard deviation in $[10, 20]$.
    \item \textbf{Brightness adjustment (p=0.3):} multiplicative factor in $[0.8, 1.2]$ with additive offset in $[-5, 5]$.
    \item \textbf{Gamma contrast (p=0.3):} gamma sampled from $[0.8, 1.2]$.
    \item \textbf{Linear contrast (p=0.1):} contrast scaling in $[0.8, 1.2]$.
    
    \item \textbf{Local intensity perturbation (p=0.3):} simplex-noise-based alpha blending with per-channel scaling in $[1.5, 2.5]$.
    
    \item \textbf{Elastic deformation:} spatial warping with $\alpha \in [15, 40]$ and $\sigma \in [5, 10]$.
\end{itemize}

\section{Reinforcement Learning Hyperparameters}
\label{sec:appendix_rl_hyperparams}

To ensure the reproducibility of our curriculum-based RL training (Ours-DS (RL)), we outline the complete hyperparameter configuration in Table~\ref{tab:rl_hyperparameters}. We optimize our policy model using Group Relative Policy Optimization (GRPO) to stabilize multi-objective reward coordination without the overhead of a separate critic network.

\begin{table}[h]
\centering
\caption{Hyperparameters used in GRPO Training.}
\label{tab:rl_hyperparameters}
\begin{tabular}{lc}
\hline
\textbf{Hyperparameter} & \textbf{Value} \\ \hline
Optimizer & AdamW \\
Learning Rate ($\eta$) & $5 \times 10^{-6}$ \\
Learning Rate Scheduler & Cosine with Warmup \\
Warmup Ratio & $0.03$ \\
Weight Decay & $0.01$ \\
\hline
Group Size ($G$) & $4$ \\
KL Coefficient ($\beta$) & $0.05$ \\
Clip $\epsilon$ (GRPO Policy Objective) & $0.2$ \\
Gradient Clipping Limit & $1.0$ \\
\hline
Max Output Length & $2048$ tokens \\
Sampling Temperature ($T$) & $0.9$ \\
Top-$p$ Sampling & $0.95$ \\
\hline
Batch Size & $16$ \\
Total Training Steps / Epochs & $1$ \\
Precision / Format & bfloat16 \\
\hline
\end{tabular}
\end{table}

As discussed in the main text, the policy rewards are computed relatively within each generated group of size $G=4$. The KL divergence penalty ($\beta = 0.05$) ensures that the generated TikZ code does not structurally degenerate or deviate catastrophically from the reference instruct model's distribution during early iterations of optimization.

\paragraph{Failure Cases.}
As shown in Figure~\ref{fig:failure_cases}, the primary failure modes involve overlapping text regions and structural misalignment of diagram elements. These errors are likely attributed to the limited size and diversity of the training dataset, which may not sufficiently capture complex sketch layouts and dense annotations. We anticipate that scaling to a larger and more diverse hand-drawn sketch dataset will improve the model's robustness and reduce these failures.

\bibliography{egbib}